# MammoRGB: Dual-View Mammogram Synthesis Using Denoising Diffusion Probabilistic Models

Authors

- Jorge Alberto Garza-Abdala, MSc [a]
- Gerardo A. Fumagal-González, BS [a]
- Daly Avendano, MD PhD [b]
- Servando Cardona, MD [b]
- Sadam Hussain, PhD [a]
- Eduardo de Avila-Armenta, MSc [a]
- Jasiel H. Toscano-Martínez, MSc [c]
- Diana S. M. Rosales Gurmendi, MSc [a]
- Alma A. Pedro-Pérez, MSc [c]
- Jose Gerardo Tamez-Pena, PhD [b]

Affiliation

a) Tecnologico de Monterrey, School of Engineering and Sciences Av. Eugenio Garza Sada 2501 Sur, Tecnológico, 64849 Monterrey, N.L.
b) Tecnologico de Monterrey, School of Medicine and Health Sciences, Av. Eugenio Garza Sada 2501 Sur, Tecnológico, 64849 Monterrey, N.L.
c) Department of Computer Science, School of Engineering, Pontificia Universidad Católica de Chile, Santiago, Chile.

Corresponding author information

Jorge Alberto Garza-Abdala

+52 5551059485

a01376671@tec.mx

Av. Eugenio Garza Sada 2501 Sur, Tecnológico, 64849 Monterrey, N.L.

**Synthetic dual-view mammograms generated by diffusion models showed structural consistency with real data, highlighting their potential to support breast cancer imaging research and AI development.**

Key points:

- Diffusion models generated anatomically consistent dual-view mammograms, with overlap metrics comparable to real datasets despite statistical differences from sample size imbalance.
- Models using the sum or difference of channels achieved the highest similarity to real data, while the zero-channel model performed the weakest.
- Artifacts remain a limitation, mainly from preprocessing, but the approach shows potential for data augmentation and training in computer-aided diagnosis.

# Abstract

Purpose

This study aims to develop and evaluate a three channel denoising diffusion probabilistic model (DDPM) for synthesizing single breast dual-view mammograms and to assess the impact of channel representations on image fidelity and cross-view consistency.

Materials and Methods

A pretrained three channel DDPM, sourced from Hugging Face, was fine-tuned on a private dataset of 11,020 screening mammograms to generate paired craniocaudal (CC) and mediolateral oblique (MLO) views. Three third-channel encodings of the CC and MLO views were evaluated: sum, absolute difference, and zero-channel. Each model produced 500 synthetic image pairs. Quantitative assessment involved breast mask segmentation using Intersection over Union (IoU) and Dice Similarity Coefficient (DSC), with distributional comparisons against 2,500 real pairs using Earth Mover's Distance (EMD) and Kolmogorov–Smirnov (KS) tests. Qualitative evaluation included a visual Turing test by a non-expert radiologist to assess cross-view consistency and artifacts.

Results

Synthetic mammograms showed IoU and DSC distributions comparable to real images, with EMD and KS values (0.020 and 0.077 respectively). Models using sum or absolute difference encodings outperformed others in IoU and DSC ($p < 0.001$), though distributions remained broadly similar. Generated CC and MLO views maintained cross-view consistency, with 6–8% of synthetic images exhibiting artifacts consistent with those in the training data.

Conclusion

Three channel DDPMs can generate realistic, anatomically consistent dual-view mammograms of a single breast. Sum and absolute difference encodings yield superior alignment with real data, supporting their use in dataset augmentation and advancing robust AI development for breast imaging.

# Introduction

Breast cancer remains a major global health concern for women, causing more than 600,000 deaths worldwide in 2022. Mortality can be reduced when the disease is detected and treated at an early stage [1]. Early detection is possible through several screening methods, with mammography being the gold standard because it is noninvasive, relatively inexpensive, and has sensitivity that increases with age [2], [3]. Mammography includes two standard views—the mediolateral oblique (MLO) and the craniocaudal (CC)—each providing complementary information.

Recent studies combining artificial intelligence (AI) with mammography have shown promise as support tools for radiologists in tasks such as early diagnosis, prognosis, lesion detection, risk assessment, and treatment evaluation [4], [5]. Moreover, using both mammographic views together has been shown to improve diagnostic accuracy compared to using a single view [6], [7]. However, publicly available datasets often suffer from class imbalance, which can lead to biased AI models [8]. Traditional augmentation methods, such as rotation, flipping, or shifting, can partially address this issue, but they only modify existing images rather than creating new ones [9].

Generative AI methods, such as Generative Adversarial Networks (GANs), Variational Autoencoders (VAEs), and diffusion models (DMs), provide an alternative. Among these, DMs have gained attention because they are more stable to train than GANs, produce high-quality samples that better represent the data distribution, and generally outperform GANs and VAEs in image generation tasks [10], [11]. Although the have been lately used to generated medical images, these efforts have been limited to the generation of a single view images, hence they can't be used to generated single breast dual view screening mammograms.

The aim of this study is to tailor and fine tune a pretrained diffusion-based model for the generation of both screening mammographic views (MLO and CC) of a single breast. Tailoring consisted in encoding the two views in the diffusion model and finding the right learning rate for dual view generation. Three strategies were evaluated: the sum of views (MLO + CC), the absolute difference between views (|MLO − CC|), and a zero channel. Hence to our knowledge, this is the first work to explore the simultaneous generation of dual-view screening mammograms

# Materials and Methods

## Dataset

This retrospective study used the private TecSalud mammography dataset, a longitudinal collection of screening images acquired between 2014 and 2019. A subset of 11,020 patients was selected, corresponding to 22,040 dual-view mammograms, including CC and MLO views [12], [13]. Only Breast Imaging Reporting and Data System (BI-RADS) category 2 cases without implants were included. The study was approved by the local institutional ethics committee (protocol number: P000542-MIRAI-MODIFICADO-CEIC-CR002).

All images' intensities were normalized to a maximum value of 1, and image resolution was set to 0.1 mm per pixel. The left CC and MLO views were mirrored to ensure consistent orientation. After that custom histogram-matching procedure was applied using the Radiological Society of North America (RSNA) dataset as a reference. The reference cumulative distribution function (CDF) was estimated from 100 randomly selected RSNA 2023 Challenge images (See Supplementary Figure S1) [14].

## Diffusion Models

Diffusion models generate synthetic images by learning to reverse a gradual noising process. During training, an image is progressively corrupted with Gaussian noise through a sequence of steps until it becomes nearly indistinguishable from random noise (the forward process). The model, typically a U-Net architecture, is then trained to predict and remove this noise at each step, effectively learning the reverse process [11].

The forward process can be written as:

$$x_t = \sqrt{\alpha_t} x_0 + \sqrt{1 - \alpha_t} \epsilon$$

where $x_0$ is the original image, $\epsilon$ is Gaussian noise, and $\alpha_t$ controls the noise level at step $t$.

During generation, the process is inverted: the model starts from pure Gaussian noise and iteratively denoises it, producing a synthetic image that follows the learned data distribution. This iterative mechanism enables diffusion models to generate samples with high fidelity and diversity while avoiding issues common in other generative models, such as mode collapse [11].

In this study, we adapted a denoising diffusion probabilistic model (DDPM) to synthesize dual-view mammograms [11]. The CC and MLO views were encoded in the red and green channels, respectively, while different strategies were tested for the blue channel (sum, difference, or zero). This design allowed the simultaneous generation of both views while maintaining anatomical consistency between them.

## Data Representation and Input Encoding

To enable the simultaneous generation of both mammographic views, each image was represented in RGB format. The red channel encoded the CC view, the green channel encoded the MLO view, and the blue channel encoded a combination of both.

Three blue-channel strategies were evaluated: the sum of the two views (Model_sum), the absolute difference between views (Model_diff), and a zero channel (Model_zero).

Figure 1 illustrates the RGB encoding scheme and the resulting input to the diffusion model. These encoded images were used directly for model fine-tuning.

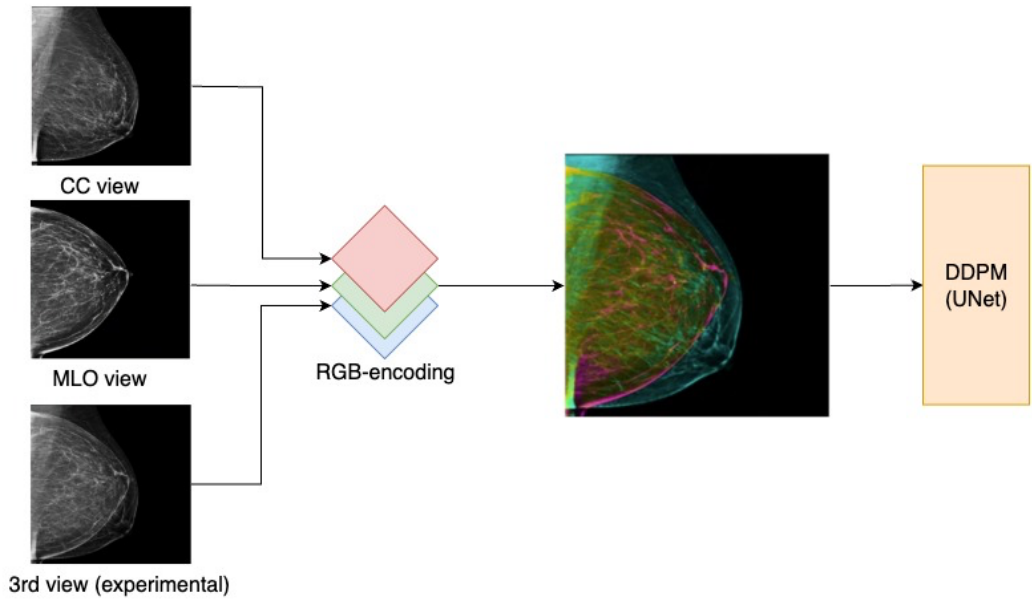

Figure 1. RGB input encoding for the DDPM. The preprocessed mammograms are encoded into an RGB image, where the red channel represents the CC view, the green channel represents the MLO view, and the blue channel encodes the 3rd view (experimental), included for testing different input variants but not used in the evaluation.

Training setup

Two main sets of experiments were conducted. The first evaluated the effect of the learning rate (LR), and the second examined the impact of the third channel in the RGB representation. All models were fine-tuned from the pretrained "*google/ddpm-celebahq-256*" checkpoint available on Hugging Face [11].

Training was performed with the following hyperparameters:

- Image size = 256x256
- Batch size = 16
- Number of epochs = 100
- LR = 1e-4 or 1e-5 (used only for the learning rate experiment), fixed at 1e-5 for all subsequent experiments

The LR of 1e-5 was selected for later experiments because it provided more stable training and better qualitative results than 1e-4.

### Three channel models comparison

To evaluate the models, we generated 500 synthetic RGB images at 256 × 256 resolution, along with the corresponding individual grayscale views. The CC view was extracted from the red channel, and the MLO view was extracted from the green channel. The third channel was defined differently across experiments (e.g., absolute difference, sum, or zeros), but it was not further analyzed in the evaluation.

Prior to metric computation, all synthetic RGB images were normalized using a 99th-percentile normalization, which scales the pixel intensities to mitigate outliers and ensure consistent contrast across images.

### Dual-view synthetic image assessment

It is expected that the generated images correspond to two views of the same breast, hence both CC and MLO views must share similar size and composition, i.e., they must be consistent. To quantitatively evaluate the size consistency between screening views, we extracted binary masks of the breast region through Otsu's thresholding method and computed the Intersection Over Union (IoU) and Dice Similarity Coefficient (DSC) between the CC and MLO masks. The real distribution of the IoU and DSC metrics were calculated using 2,500 randomly chosen real mammogram. To evaluate the similarity between real and synthetic mammograms, we compared the real distributions of IOU and DICE metrics *vs.* the distribution of generated images. The Earth Mover's Distance (EMD) was used to compare the distributions, and the statistical significance of the similarity was computed using and the Kolmogorov-Smirnov (KS). EMD measures the minimal effort required to transform one distribution into another while KS quantifies the maximum deviation between two cumulative distributions [15], [16].

Finally, a visual inspection was performed by a non-expert to look for artifacts and qualitatively evaluate anatomical coherence between both views: e.g., Breast size, breast density.

# Results

We successfully generated both mammographic views (CC and MLO) using an RGB-based representation. Figure 2 shows the generations obtained from both models at 10, 20, 50, and 70 training epochs. The top row presents results from the model trained with a LR of 1e-4, while the bottom row shows results from the model trained with LR = 1e-5. The first model required nearly 70 epochs to produce visually plausible mammograms, whereas the second model generated anatomically consistent structures as early as 20 epochs. Additionally, the second model displayed clearer breast boundaries and fewer artifacts throughout training. Based on these observations, all subsequent experiments and evaluations were conducted using the models trained with LR = 1e-5.

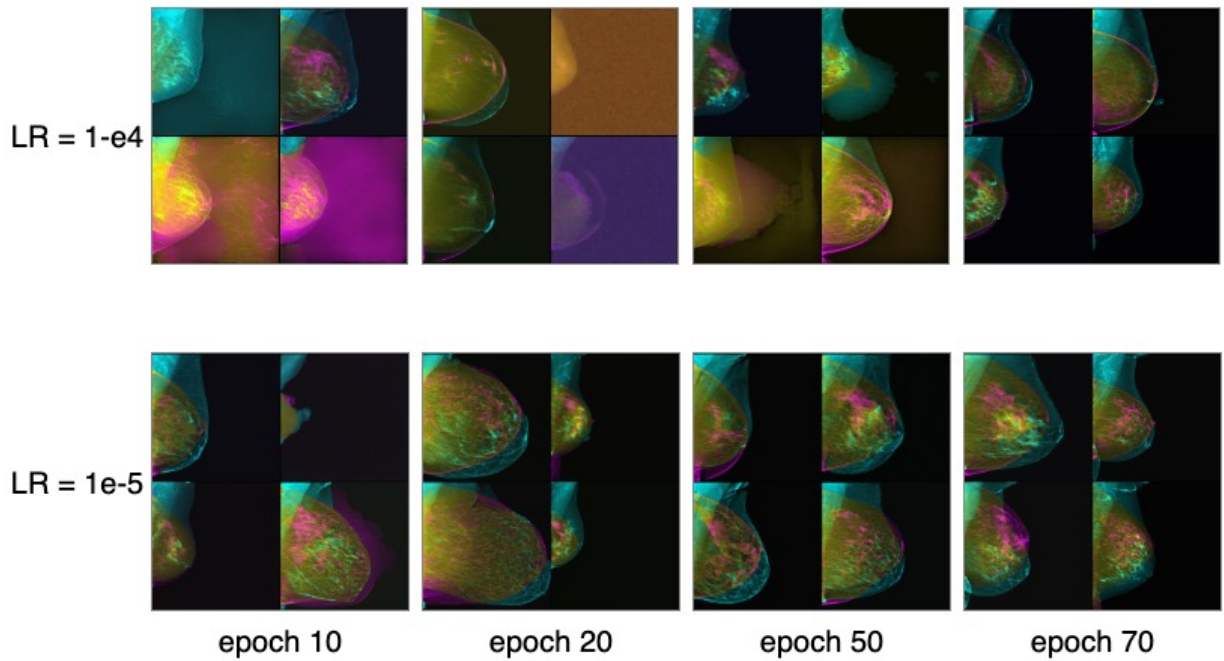

Figure 2. Synthetic images at different epochs. The top row is for the model trained with a learning rate of 1e-4, while the bottom row is for the model trained with a learning rate of 1e-5. From left to right: set of four images generated at epochs 10, 20, 50, and 70.

The visual inspection confirmed that all models trained with LR = 1e-5 maintained anatomical consistency between CC and MLO views (Figure 3). The incidence of cross-view artifacts was high and visually present on 98% images (see Supplementary Figure S3). The frequency of major artifacts varied across models: approximately 6% of paired images (30/500) for Model_sum, 6.2% (31/500) for Model_diff, and 7.6% (38/500) for Model_zeros, and corresponding to artifacts present in the training set.

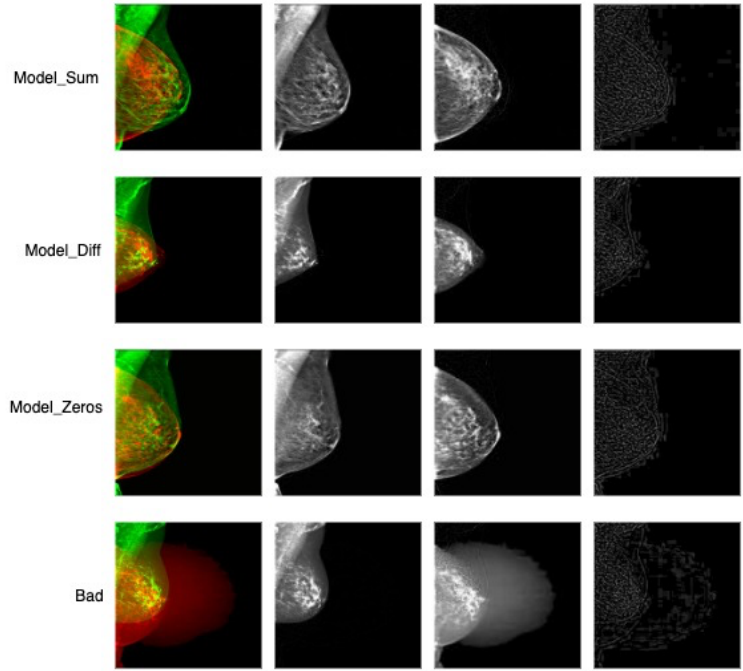

Figure 3. Examples of synthetic mammograms generated by the proposed models. Each row represents a different randomly selected test sample, with the last row illustrating a case with major artifacts. From left to right: the synthetic RGB mammogram normalized to 99%, the CC view (red channel), the MLO view (green channel), and the third channel. The third channel varies across experiments (sum, absolute difference, or zeros), and its brightness was enhanced to highlight the differences.

Figure 4 presents a sample of the IoU and DSC results. The distributions of these metrics across all models are illustrated as violin plots in Figure 5, while detailed descriptive statistics—including mean, standard deviation, first quartile, median, third quartile, maximum, and interquartile range—are provided in Table 1.

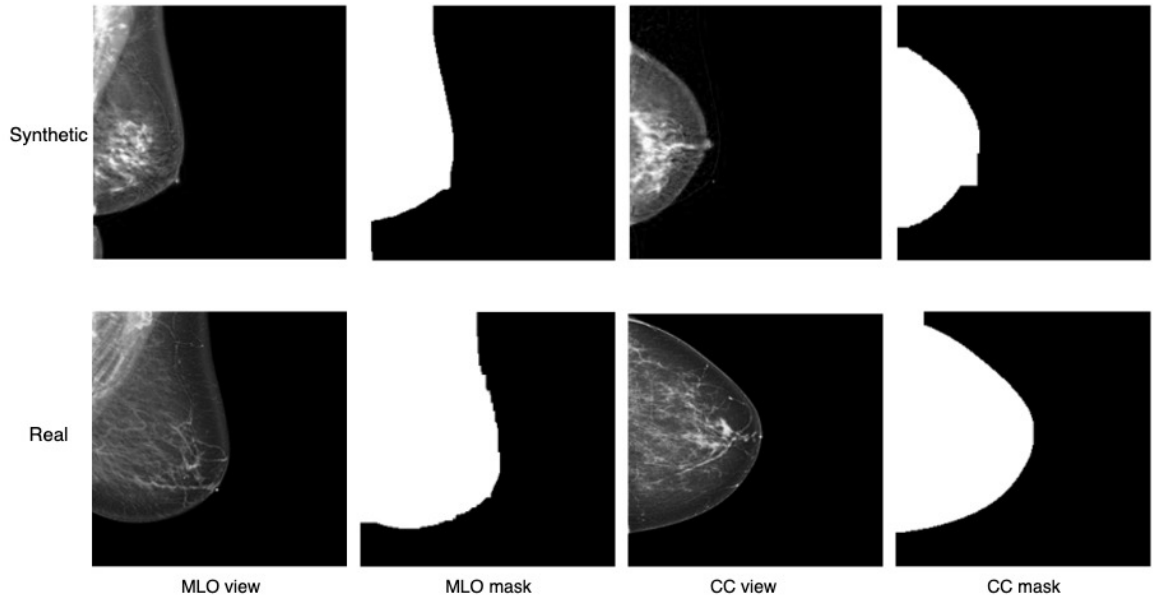

Figure 4. Comparison of real and synthetic masks through Otsu's thresholding method. The top row is for synthetic images, and the bottom row is for real images. From left to right: MLO view, MLO mask, CC view, CC's mask.

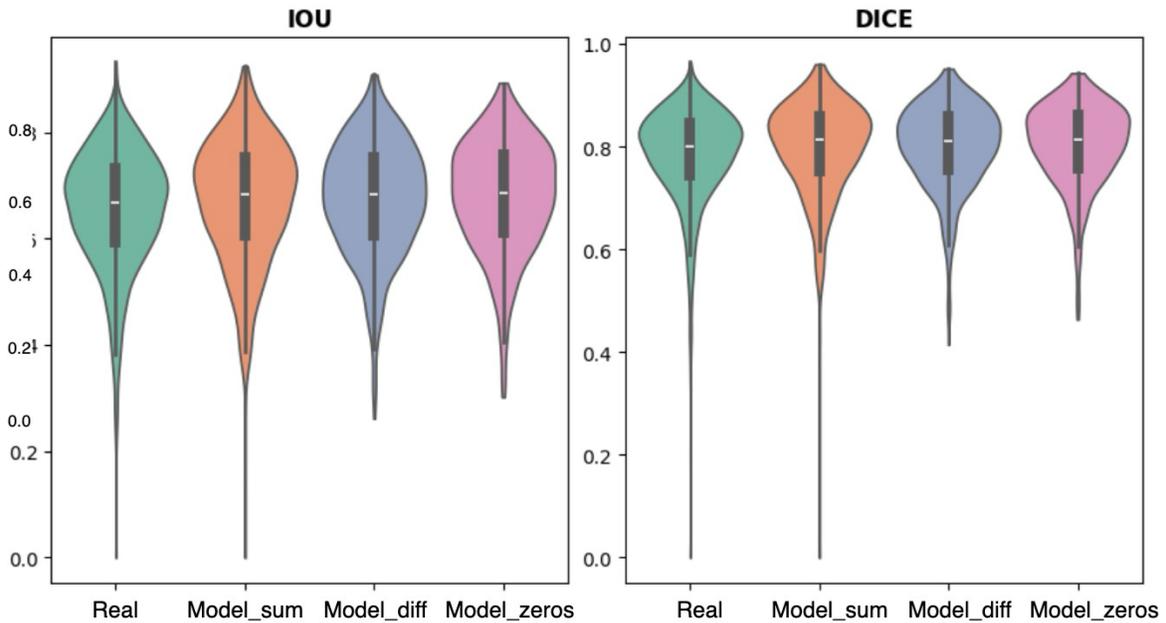

Figure 5. Distributional comparison of metrics for synthetic and real mammogram masks. From left to right, IOU and DICE violin plots for each dataset or synthetic image configuration. The five violin plots in each square correspond to Real (teal), Model_sum (orange), Model_diff (blue) and Model_zeros (pink).

Table 2 summarizes the comparison of IoU and DSC across the three synthetic datasets relative to the real dataset. For both metrics, small but statistically significant differences were observed for Model_sum and Model_diff according to the KS test. However, EMD values remained low, suggesting that the synthetic images broadly replicate the structural characteristics of real mammograms. Overall, both encodings showed the agreement with real mammograms, both qualitatively and quantitatively.

# Discussion

In this work, we demonstrated that the simultaneous generation of dual-view mammograms is possible by encoding original mammograms as three channel images using DDPM. Although mammogram synthesis has been explored in previous studies, to our knowledge, this is the first demonstration of simultaneous synthesis of both CC and MLO views [17], [18], [19], [20].

Our results showed that the proposed models achieved mean and standard deviation values of IoU and DSC that were close to those of real images. This indicates that the generated views maintained consistent size and alignment. In addition, KS and EMD tests suggested that the distributions of these metrics for synthetic datasets were comparable to those of real data.

All models successfully generated dual-view mammograms, but the choice of the third channel had a clear impact on performance. Models using sum or difference information (Model_sum and Model_diff) achieved the best results, showing higher anatomical consistency. In contrast, the zero-channel model (Model_zero) performed the worst, with the highest EMD and KS values ($p < 0.001$).

The visual evaluation indicated that the artifacts were present in the training images step rather than visual hallucinations of the generative models themselves (see Supplementary Figure S2). Alternative preprocessing strategies could reduce or even eliminate these issues. Moreover, the presence of artifacts in generated images may reflect the potential for simulating tumor-like structures: if training data contain no artifacts, synthetic images are unlikely to show them

Compared with prior work, our approach makes a distinct contribution by generating paired CC and MLO views with patient-specific correspondence. Previous studies have mainly focused on single-view synthesis. For example, Montoya-del-Angel et al. introduced MAM-E, a stable diffusion–based method for generating full-field mammograms and performing lesion inpainting through prompts and mask control [17]. Sutjiadi et al. fine-tuned a pretrained DDPM on the INbreast and MIAS datasets, while Meng et al. proposed a DDPM-based approach that integrates breast region segmentation and prior knowledge [18], [21]. Joseph et al. developed a GAN-based model capable of producing class-labeled mammograms with realistic tissue preservation [19]. Yamazaki et al. explored generating one view from another using a GAN-based image-to-image model, but their work did not address simultaneous synthesis [20]. While these methods show strong potential for clinical applications, none have addressed the challenge of producing anatomically consistent dual views.

## Limitations and future work

This study has several limitations. First, it focused exclusively on generating dual-view synthetic mammograms and did not evaluate their effect into downstream deep learning pipelines. Second, the assessment of anatomical consistency was limited to a masking-based method, without exploring other internal anatomical metrics. Third, no radiologist participated in the visual inspection, as the synthetic mammograms still present minor artifacts that make them unsuitable for clinical-level evaluation at this stage. Nevertheless, our findings provide evidence that generating dual-view mammograms is feasible, establishing a foundation for further investigation.

Future work will aim to address these limitations. We plan to explore the use of stable diffusion pipelines for conditional generation to improve anatomical fidelity and reduce artifacts. Another important direction is testing the utility of paired synthetic images in deep learning tasks, such as classification or detection, to assess their potential for data augmentation. Additionally, future studies will investigate new quantitative metrics for evaluating internal anatomical consistency across views. Finally, we will examine the generalizability of our approach by applying it to different generative models and datasets.

## Conclusions

This study presents an approach for generating dual-view mammograms by encoding both views in three channel images, allowing the model to capture cross-view anatomical relationships and improve single-view mammogram synthesis.

Our evaluation showed that the proposed models achieved IoU and DSC scores comparable to those of real datasets. However, statistical tests revealed differences likely due to the imbalance in sample sizes. The design of the third channel also influenced performance: models using sum or difference information showed the best alignment with real data, while the zero-channel performed the weakest.

Although artifacts remain a limitation—mainly related to preprocessing—our findings indicate that diffusion models can be adapted to generate anatomically consistent dual-view mammograms. This opens the possibility of using synthetic image pairs for data augmentation and training in computer-aided diagnosis.

Future work should focus on applying this approach to other generative AI models, such as Stable Diffusion, to further reduce artifacts and enhance clinical applicability. It will also be important to validate the method across different medical imaging datasets.

## Acknowledgments

This research was supported by the Secretaría de Ciencia, Humanidades, Tecnología e Innovación (Secihti), with cloud computing resources provided through Microsoft's AI for Good Lab.

| | | | IoU | | | | | | | |
|---|---|---|---|---|---|---|---|---|---|---|
| Dataset | Count | Mean | Mean difference (vs real) | Std | Min | Q1 | Median | Q3 | Max | IQR |
| Real | 2500 | 0.654 | N/A | 0.113 | 0 | 0.592 | 0.668 | 0.733 | 0.932 | 0.141 |
| Model_sum | 500 | 0.670 | **0.016** | 0.117 | 0 | 0.605 | 0.684 | 0.754 | 0.924 | 0.149 |
| Model_diff | 500 | 0.674 | 0.020 | 0.107 | 0.261 | 0.605 | 0.682 | 0.751 | 0.908 | 0.146 |
| Model_zeros | 500 | 0.678 | 0.024 | 0.105 | 0.302 | 0.610 | 0.686 | 0.757 | 0.893 | 0.147 |
| | | | DSC | | | | | | | |
| Dataset | Count | Mean | Mean difference (vs real) | Std | Min | Q1 | Median | Q3 | Max | IQR |
| Real | 2500 | 0.784 | N/A | 0.092 | 0 | 0.743 | 0.801 | 0.846 | 0.964 | 0.102 |
| Model_sum | 500 | 0.795 | **0.011** | 0.093 | 0 | 0.753 | 0.812 | 0.859 | 0.960 | 0.105 |
| Model_diff | 500 | 0.800 | 0.016 | 0.081 | 0.414 | 0.754 | 0.811 | 0.858 | 0.925 | 0.103 |
| Model_zeros | 500 | 0.803 | 0.019 | 0.079 | 0.463 | 0.757 | 0.813 | 0.862 | 0.943 | 0.104 |

Table 1. Descriptive statistics of IoU, DSC, and pixel-wise intersection metrics for real and synthetic mammogram masks. The table is divided into three sections, one for each metric. For each dataset, the reported values include Count, Mean, Standard Deviation (Std), Minimum (Min), First Quartile (Q1), Median, Third Quartile (Q3), Maximum (Max), and Interquartile Range (IQR). The real dataset includes 2,500 samples, while each synthetic dataset includes 500 samples.

| IoU | | | |
|---|---|---|---|
| Dataset | EMD | KS D | p-value |
| Model_sum | 0.016 | 0.097 | * |
| Model_diff | **0.020** | **0.077** | ** |
| Model_zeros | 0.024 | 0.100 | * |
| **DSC** | | | |
| Dataset | EMD | KS D | p-value |
| Model_sum | **0.012** | 0.097 | * |
| Model_diff | 0.015 | **0.077** | ** |
| Model_zeros | 0.019 | 0.100 | * |

Table 2. Statistical comparison of IoU, DSC, and Intersection metrics between real and synthetic mammograms using Earth Mover's Distance (EMD) and Kolmogorov-Smirnov (KS) tests. Lower EMD and KS D values indicate higher similarity. ∗ = p<0.001, ∗∗ = 0.005<p<0.05